\documentclass[final]{article}
\usepackage{neurips_2019}
\usepackage[utf8]{inputenc} % allow utf-8 input
\usepackage[T1]{fontenc}    % use 8-bit T1 fonts
\usepackage{times}
\usepackage{hyperref}       % hyperlinks
\usepackage{url}            % simple URL typesetting
\usepackage{nicefrac}       % compact symbols for 1/2, etc.
\usepackage{amsmath,amsfonts,amssymb}
\usepackage{graphicx}
\usepackage[dvipsnames]{xcolor}
\usepackage{booktabs}
\usepackage{xcolor}
\usepackage{placeins}
\graphicspath{{}{figures/}}

\def\erfc{{\mathrm{erfc}}}
\def\erfcm1{{\mathrm{erfc}^{\!-1}}}
\def\changed#1{\textcolor{purple}{#1}}
\def\changed#1{{#1}} % comment this out to show changes in purple

\newsavebox{\savepar}

\newif\ifnotanonymous\notanonymousfalse
\makeatletter
\if@neuripsfinal\notanonymoustrue\fi
\if@preprint\notanonymoustrue\fi
\makeatother

\newif\ifsupplementary
\supplementarytrue

\title{Cold Case: the Lost MNIST Digits}

\author{
   Chhavi Yadav\\
   New York University\\
   New York, NY\\
   \texttt{chhavi@nyu.edu}
   \And
   L\'{e}on Bottou\\
   Facebook AI Research\\
   and New York University\\
   New York, NY\\
   \texttt{leon@bottou.org}
 }

\begin{document}
\maketitle

\begin{abstract}

%[LEON'S VERSION]

Although the popular MNIST dataset \citep{mnist} is derived
from the NIST database \citep{nist-sd19}, the
precise processing steps for this derivation have been lost to time. We propose
a reconstruction that is accurate enough to
serve as a replacement for the MNIST dataset, with insignificant changes in accuracy. 
We trace each MNIST digit to its NIST source and its rich 
metadata such as writer identifier, partition identifier, etc.  
We also reconstruct the complete MNIST test set with 60,000 
samples instead of the usual 10,000. Since the balance 50,000 were 
never distributed, they can be used to investigate the impact of twenty-five years
of MNIST experiments on the reported testing performances.
Our limited results unambiguously confirm the trends observed by 
\citet{recht2018cifar, pmlr-v97-recht19a}:
although the misclassification rates are slightly off, 
classifier ordering and model selection remain broadly reliable.
We attribute this phenomenon to the pairing benefits
of comparing classifiers on the same digits.

%[CHHAVI'S VERSION]

% The MNIST dataset \citep{mnist,bottou-cortes-94} is a popular machine learning benchmark. It consists of 60,000 training and 10,000 testing images presently. Going back in time, the authors of MNIST had created two sets of 60,000 unprocessed images each. Out of these two, the testing set was downsampled to 10,000. We rediscover the lost 50,000 testing images in our work.
% We reconstruct both the training and testing sets from NIST and preprocess them. We call this the Quasi-MNIST (Q-MNIST) dataset.  We also experimentally verify that the reconstruction is precise. Hence, now we have a new and larger testing set with almost the same distribution as the MNIST testing set. This enables us to check for potential `testing set rot' as one can interpret the succession of papers using MNIST over 20 years, as a learning system that overfits on the MNIST testing set. Accuracies of selected machine learning models dropped when tested on Q-MNIST testing set. However, relative ordering of models was maintained. This suggests that machine learning benchmarks do remain reliable after many years.
\end{abstract}

% ========================================
\section{Introduction}

The MNIST dataset \citep{mnist,bottou-cortes-94}
has been used as a standard machine learning benchmark for more than
twenty years. During the last decade, many researchers have expressed the opinion that this dataset has been overused.  In particular, the small size of its test set, merely 10,000 samples, has been a
cause of concern. Hundreds of publications report increasingly good
performance on this same test set. Did they overfit the test
set? Can we trust any new conclusion drawn on this dataset?  How
quickly do machine learning datasets become useless?

The first partitions of the large NIST handwritten character
collection \citep{nist-sd19} had been released one year earlier, with a
training set written by 2000 Census Bureau employees and a
substantially more challenging test set written by 500 high school
students. One of the objectives of LeCun, Cortes, and
Burges was to create a dataset with similarly distributed
training and test sets.  The process they describe produces two
sets of 60,000 samples. The test set was then downsampled to only
10,000 samples, possibly because manipulating such a dataset with the
computers of the times could be annoyingly slow. The remaining 50,000
test samples have since been lost.

The initial purpose of this work was to recreate the MNIST
preprocessing algorithms in order to trace back each MNIST digit to
its original writer in NIST. This reconstruction was first based on the available information and then considerably improved by iterative refinements. Section~\ref{sec:reconstruction} describes this process and measures how closely our reconstructed samples match the official MNIST
samples. The reconstructed training set contains 60,000 images
matching each of the MNIST training images.  Similarly, the first
10,000 images of the reconstructed test set match each of the MNIST
test set images. The next 50,000 images are a reconstruction of
the 50,000 lost MNIST test images.\footnote{\relax
\ifnotanonymous
Code and data are available at \url{https://github.com/facebookresearch/qmnist}.
\else
We of course intend to publish both the reconstruction code and 
the reconstructed dataset.
\fi
}

In the same spirit as~\citep{recht2018cifar,pmlr-v97-recht19a},
the rediscovery of the 50,000 lost MNIST test digits provides an
opportunity to quantify the degradation of the official MNIST test
set over a quarter-century of experimental research. 
Section~\ref{sec:genex} compares and discusses the performances
of well known algorithms measured on the original MNIST test
samples, on their reconstructions, and on the reconstructions of the
50,000 lost test samples. Our results provide a well controlled 
confirmation of the trends identified 
by~\citet{recht2018cifar,pmlr-v97-recht19a} 
on a different dataset.

\begin{figure}[t]
  \begin{lrbox}{\savepar}
    \begin{minipage}{.8\linewidth}
      \usefont{OT1}{pbk}{m}{n}\small
      \definecolor{mygray}{gray}{0.25}\color{mygray}
      The original NIST test contains 58,527 digit images written by
      500 different writers.  In contrast to the training set, where
      blocks of data from each writer appeared in sequence, the data in
      the NIST test set is scrambled. Writer identities for the test
      set is available and we used this information to unscramble the
      writers. We then split this NIST test set in two: characters
      written by the first 250 writers went into our new training
      set. The remaining 250 writers were placed in our test set. Thus
      we had two sets with nearly 30,000 examples each.
      \par\smallskip
      \noindent\mbox{\quad}
      The new training set was completed with
      enough samples from the old NIST training set, starting at
      pattern \#0, to make a full set of 60,000 training
      patterns. Similarly, the new test set was completed with old
      training examples starting at pattern \#35,000 to make a full
      set with 60,000 test patterns. All the images were size
      normalized to fit in a 20 x 20 pixel box, and were then centered
      to fit in a 28 x 28 image using center of gravity. Grayscale
      pixel values were used to reduce the effects of aliasing. These
      are the training and test sets used in the benchmarks described
      in this paper. In this paper, we will call them the MNIST data.
    \end{minipage}
  \end{lrbox}
  \centering
  \fbox{\usebox{\savepar}}
  \caption{\label{fig:mnist} The two paragraphs of
    \citet{bottou-cortes-94} describing the MNIST preprocessing. The
    hsf4 partition of the NIST dataset, that is, the original test
    set, contains in fact 58,646 digits.}
\end{figure}

% ========================================
\section{Recreating MNIST}
\label{sec:reconstruction}

Recreating the algorithms that were used to construct the MNIST
dataset is a challenging task. Figure~\ref{fig:mnist} shows the two
paragraphs that describe this process
in~\citep{bottou-cortes-94}. Although this was the first paper
mentioning MNIST,
the creation of the dataset predates this
benchmarking effort by several months.\relax
{\ifnotanonymous\footnote{When LB joined this
  effort during the summer 1994, the MNIST dataset was already ready.}\fi}
Curiously, this description incorrectly reports that the number of digits
in the hsf4 partition, that is, the original NIST testing set, 
as 58,527 instead of 58,646.\relax
\footnote{The same description also appears in \citep{mnist,lecun-98h}. 
  These more recent texts incorrectly use the names
  SD1 and SD3 to denote the original NIST test and training
  sets. And additional sentence explains that only a subset of 10,000
  test images was used or made available,
  ``\emph{5000 from SD1 and 5000 from SD3}.''}

These two paragraphs give a relatively precise recipe for selecting the
60,000 digits that compose the MNIST training set. Alas, applying
this recipe produces a set that contains one more zero and one
less eight than the actual MNIST training set. Although they
do not match, these class distributions are too close to make it plausible
that 119 digits were really missing from the hsf4 partition. 

The description of the image processing steps is much less precise.
How are the 128x128 binary NIST images cropped? Which heuristics, if
any, are used to disregard noisy pixels that do not belong to the
digits themselves? How are rectangular crops centered in
a square image? How are these square images resampled to 20x20
gray level images? How are the coordinates of the center of gravity
rounded for the final centering step?

\subsection{An iterative process}

Our initial reconstruction algorithms were informed by the existing
description and, crucially, by our knowledge of a mysterious resampling
algorithm found in ancient parts of the Lush codebase: instead of
using a bilinear or bicubic interpolation, this code computes the
exact overlap of the input and output image pixels.\relax
\footnote{See \url{https://tinyurl.com/y5z7qtcg}.}

Although our first reconstructed dataset, dubbed QMNISTv1, behaves very much
like MNIST in machine learning experiments, its digit images could not
be reliably matched to the actual MNIST digits. In fact, because many
digits have similar shapes, we must rely on subtler details such as
the anti-aliasing pixel patterns.  It was however possible to identify
a few matches. For instance we found that the lightest zero in the
QMNIST training set matches the lightest zero in the MNIST training
set. We were able to reproduce their antialiasing patterns by
fine-tuning the initial centering and resampling algorithms, leading
to QMNISTv2.

We then found that the smallest $L_2$ distance between MNIST digits
and jittered QMNIST digits was a reliable match indicator.  Running
the Hungarian assignment algorithm on the two training sets gave good
matches for most digits. A careful inspection of the worst matches
allowed us to further tune the cropping algorithms, and to discover,
for instance, that the extra zero in the reconstructed training set was
in fact a duplicate digit that the MNIST creators had identified and
removed. The ability to obtain reliable matches allowed us to iterate
much faster and explore more aspects the image processing algorithm
space, leading to QMNISTv3, v4, and v5.
Note that all this tuning was achieved by
matching training set images only.

\begin{figure}[t]
  \centering
  \includegraphics[width=.67\linewidth]{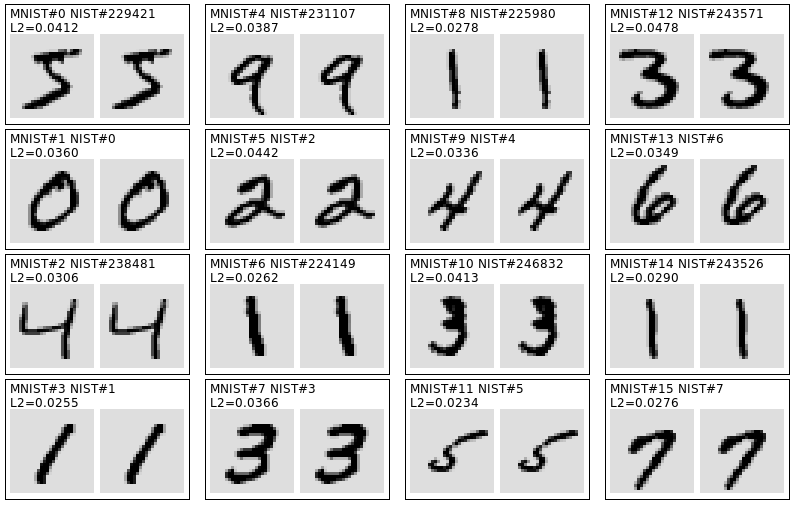}
  \par\smallskip
  \begin{lrbox}{\savepar}
    \begin{tabular}{r@{}c}
      \parbox{7em}{\small\sf Magnification:\\MNIST\,\#0\\NIST\,\#229421} &
      \parbox{.435\linewidth}{\includegraphics[width=\linewidth]{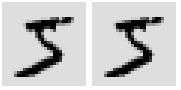}}
    \end{tabular}
  \end{lrbox}
  \fbox{\usebox{\savepar}}
  \caption{\label{fig:train20} Side-by-side display of the first
    sixteen digits in the MNIST and QMNIST training set. The magnified
    view of the first one illustrates the correct reconstruction of the
    antialiased pixels.}
\end{figure}

This seemingly pointless quest for an exact reconstruction was
surprisingly addictive. Supposedly urgent tasks could be indefinitely
delayed with this important procrastination pretext. Since all good
things must come to an end, we eventually had to freeze one of these
datasets and call it QMNIST.

\begin{table}[t]
  \centering
  \caption{\label{tbl:qmnist} Quartiles of the jittered distances
    between matching MNIST and QMNIST training digit images with
    pixels in range $0\dots255$. A $L_2$ distance of $255$ would
    indicate a one pixel difference. The $L_\infty$ distance represents
    the largest absolute difference between image pixels.}
  \smallskip
  \begin{tabular}{lccccc}
    \toprule
    & \bf Min & \bf 25\% & \bf Med & \bf 75\% & \bf Max \\
    \midrule
    Jittered $L_2$ distance & 0 & 7.1 & 8.7 & 10.5 & 17.3 \\
    Jittered $L_\infty$ distance & 0 & 1 & 1 & 1 & 3 \\
    \midrule
  \end{tabular}
  \medskip
  \caption{\label{tbl:jitter} Count of training samples for which the
    MNIST and QMNIST images align best without translation or with a
    $\pm1$ pixel translation.}
  \smallskip
  \begin{tabular}{lcc}
    \toprule
    Jitter & $\mathbf{0}$ \bf pixels & \bf $\mathbf{\pm1}$ \bf pixels\\
    \midrule
    Number of matches & 59853 & 147 \\
    \midrule
  \end{tabular}
  \smallskip
\end{table}

\subsection{Evaluating the reconstruction quality}
\label{eval}
Although the QMNIST reconstructions are closer to the MNIST images
than we had envisioned, they remain imperfect.

Table~\ref{tbl:jitter} indicates that about $0.25\%$ of the QMNIST
training set images are shifted by one pixel relative to
their MNIST counterpart. This occurs when the center of gravity
computed during the last centering step (see Figure~\ref{fig:mnist})
is very close to a pixel boundary. Because the image reconstruction
is imperfect, the reconstructed center of gravity sometimes lands
on the other side of the pixel boundary, and the alignment code
shifts the image by a whole pixel.

Table~\ref{tbl:qmnist} gives the quartiles of the $L_2$ distance and
$L_\infty$ distances between the MNIST and QMNIST images, after
accounting for these occasional single pixel shifts. An $L_2$ distance
of $255$ would indicate a full pixel of difference. The $L_\infty$
distance represents the largest difference between image pixels,
expressed as integers in range $0\dots 255$.

\begin{table}
  %%% should we move this one to the end?
  \centering
  \caption{\label{tbl:lenet5} Misclassification rates of a Lenet5
    convolutional network trained on both the MNIST and QMNIST
    training sets and tested on the MNIST test set, on the 10K QMNIST
    testing examples matching the MNIST testing set, and on the
    50k remaining QMNIST testing examples.}
  \smallskip
  \begin{tabular}{lccc}
    \toprule
    \bf Test on & \bf MNIST & \bf QMNIST10K & \bf QMNIST50K \\
    \midrule
    Train on MNIST & $0.82\%$ ($\pm 0.2\%$) & $0.81\%$ ($\pm 0.2\%$) & $1.08\%$ ($\pm 0.1\%$)\\
    Train on QMNIST & $0.81\%$ ($\pm 0.2\%$) & $0.80\%$ ($\pm 0.2\%$) & $1.08\%$ ($\pm 0.1\%$) \\
    \midrule
  \end{tabular}
\end{table}

In order to further verify the reconstruction quality, we trained a
variant of the Lenet5 network described by \citet{lecun-98h}.  Its
original implementation is still available as a demonstration in the
Lush codebase. Lush \citep{lush} descends from the SN neural network
software \citep{bottou-lecun-88} and from its AT\&T Bell Laboratories
variants developped in the nineties. This particular variant of Lenet5
omits the final Euclidean layer described in~\citep{lecun-98h} without
incurring a performance penalty. Following the pattern set by the
original implementation, the training protocol consists of three sets
of 10 epochs with global stepsizes $10^{-4}$, $10^{-5}$, and $10^{-6}$. Each set
starts with estimating the diagonal of the Hessian. Per-weight
stepsizes are then computed by dividing the global stepsize by the
estimated curvature plus 0.02.  Table~\ref{tbl:lenet5} reports
insignificant differences when one trains with the MNIST or QMNIST
training set or test with MNIST test set or the matching part of
the QMNIST test set. On the other hand, we observe a more
substantial difference when testing on the remaining part of the
QMNIST test set, that is, the reconstructions of the lost MNIST
test digits.  Such discrepancies will be discussed more precisely
in Section~\ref{sec:genex}.

\subsection{MNIST trivia}

\begin{figure}
  \centering
  \includegraphics[width=.8\linewidth]{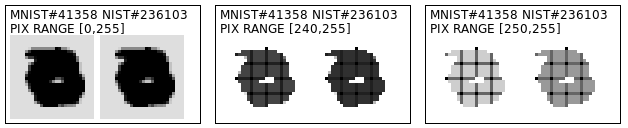}
  \caption{\label{fig:waves} We have reproduced a defect of the
    original resampling code that creates low amplitude periodic
    patterns in the dark areas of thick characters.}
  \bigskip
  \includegraphics[width=.5\linewidth]{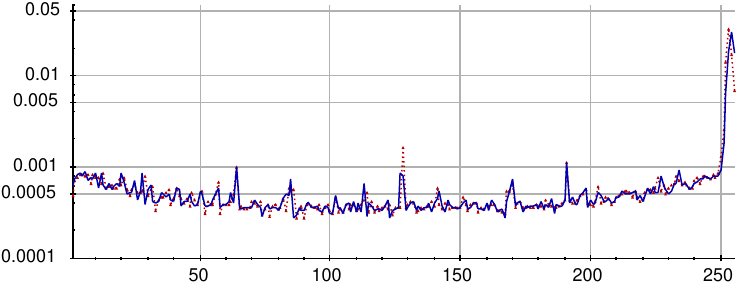}
  \caption{\label{fig:pixels} Histogram of pixel values
    in range 1-255 in the MNIST (red dots) and QMNIST (blue line)
    training set. Logarithmic scale.}
  \bigskip
  \includegraphics[width=.45\linewidth,scale=.3]{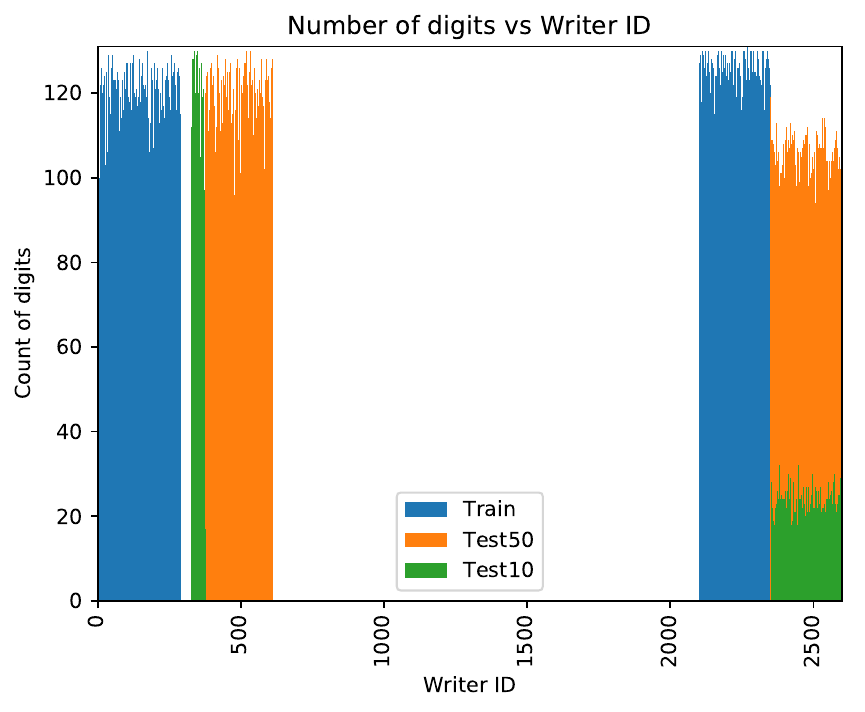}
  \caption{\label{fig:histogram} Histogram of Writer IDs and Number of digits written by the writer in MNIST Train, MNIST Test 10K and QMNIST Test 50K sets.
  }
\end{figure}

The reconstruction effort allowed us to uncover a lot
of previously unreported facts about MNIST.
\begin{enumerate}
\item
  There are exactly three duplicate digits in the entire NIST handwritten
  character collection. Only one of them falls in the segments 
  used to generate MNIST but was removed by the MNIST authors.
\item
  The first 5001 images of the MNIST test set seem randomly picked from
  those written by writers \#2350-\#2599, all high school students.
  The next 4999 images are the consecutive NIST images
  \#35,000-\#39,998, in this order, written by only 48
  Census Bureau employees, writers \#326-\#373, as shown in Figure~\ref{fig:histogram}. Although this small number
  could make us fear for statistical significance, these
  comparatively very clean images contribute little to
  the total test error.
\item
  Even-numbered images among the 58,100 first MNIST training
  set samples exactly match the digits written by writers \#2100-\#2349, all high
  school students, in random order. The remaining images are the NIST
  images \#0 to \#30949 in that order. The beginning of this sequence
  is visible in Figure~\ref{fig:train20}. Therefore, half
  of the images found in a typical minibatch of consecutive MNIST
  training images are likely to have been written by the same
  writer. We can only recommend shuffling the training set before
  assembling the minibatches.
\item
  There is a rounding error in the final centering of the 28x28 MNIST
  images. The average center of mass of a MNIST digits is in fact
  located half a pixel away from the geometrical center of the
  image. This is important because training on correctly centered
  images yields substantially worse performance 
  on the standard MNIST testing set.
\item
  A slight defect in the MNIST resampling code generates low amplitude
  periodic patterns in the dark areas of thick characters. These
  patterns, illustrated in Figure~\ref{fig:waves}, can be traced to a 0.99
  fudge factor that is still visible in the Lush legacy
  code.\footnote{See \url{https://tinyurl.com/y5z7abyt}} 
  Since the period of these patterns depend on the sizes of the input
  images passed to the resampling code, we were able to
  determine that the small NIST images were not upsampled
  by directly calling the resampling code, but by first doubling
  their resolution, then downsampling to size 20x20.
\item
  Converting the continuous-valued pixels of the subsampled images into
  integer-valued pixels is delicate. Our code linearly maps the range
  observed in each image to the interval [0.0,255.0], rounding
  to the closest integer. Comparing the pixel histograms (see
  Figure~\ref{fig:pixels}) reveals that MNIST has substantially more
  pixels with value 128 and less pixels with value 255.  We could not
  think of a plausibly simple algorithm compatible with this
  observation.
\end{enumerate}

% ========================================
\section{Generalization Experiments}
\label{sec:genex}

This section takes advantage of the reconstruction of the lost
50,000 testing samples to revisit some MNIST performance results 
reported during the last twenty-five years.
\citet{recht2018cifar,pmlr-v97-recht19a} perform a similar
study on the CIFAR10 and ImageNet datasets and identify
very interesting trends. However they also explain that they
cannot fully ascertain how closely the distribution of the reconstructed 
dataset matches the distribution of the original dataset, 
raising the possibility of the reconstructed dataset 
being substantially harder than the original.
Because the published MNIST test set was subsampled from a larger
set, we have a much tighter control of the data distribution
and can confidently confirm their findings.

Because the MNIST testing error rates are usually low, we
start with a careful discussion of the computation of
confidence intervals and of the statistical significance
of error comparisons in the context of repeated experiments.
We then report on MNIST results for several methods:
k-nearest neightbors (KNN), support vector machines (SVM),
multilayer perceptrons (MLP), and several flavors of convolutional networks (CNN).

\subsection{About confidence intervals}
\label{confint}
\changed{Since we want to know whether the actual performance of a learning system
differs from the performance estimated using an overused testing set with 
run-of-the-mill confidence intervals, all confidence intervals reported in this work}
were obtained using the classic Wald method: when we observe $n_1$ misclassifications
out of $n$ independent samples, the error rate $\nu=n_1/n$ is reported
with confidence $1{-}\eta$ as
\begin{equation}
  \label{eq:wald}
  \nu ~\pm~ z \sqrt{\frac{\nu(1-\nu)}{n}}~,
\end{equation}
where $z=\sqrt{2}\:\erfcm1(\eta)$ is approximately equal
to 2 for a 95\% confidence interval.  For instance, an error rate
close to $1.0\%$ measured on the usual 10,000 test example is
reported as a $1\%\pm0.2\%$ error rate, that is, $100\pm20$
misclassifications. This approach is widely used despite the fact that it 
only holds for a single use of the testing set \changed{and that it relies 
on an imperfect central limit approximation.}

\changed{The simplest way to account for repeated uses of the testing set
is the Bonferroni correction~\citep{bonferroni-1936}, 
that is, dividing $\eta$ by the number $K$ of potential experiments,
\emph{simultaneously defined before performing any measurement}. Although
relaxing this simultaneity constraint progressively requires all the 
apparatus of statistical learning theory~\citep[\S6.3]{vapnik-82},
the correction still takes the form of a divisor~$K$ applied to confidence level $\eta$.
Because of the asymptotic properties of the $\erfc$ function,
the width of the actual confidence intervals essentially grows 
like~$\log(K)$.}

\changed{In order to complete this picture, one also needs
to take into account the \emph{benefits} of using the same testing set.}
Ordinary confidence intervals are  overly pessimistic when we merely
want to know whether a first classifier with error rate $\nu_1=n_1/n$
is worse than a second classifier with error rate $\nu_2=n_2/n$.
Because these error rates are measured on the same test samples,
we can instead rely on a pairing argument: the first classifier
can be considered worse with confidence~$1{-}\eta$ when
\begin{equation}
  \label{eq:paired}
  \nu_1 - \nu_2 ~=~ \frac{n_{12}-n_{21}}{n} ~~\geq ~~ z \frac{\sqrt{n_{12}+n_{21}}}{n}~,
\end{equation}
where $n_{12}$ represents the count of examples misclassified by the
first classifier but not the second classifier, $n_{21}$ is the
converse, and $z=\sqrt{2}\:\erfcm1(2\eta)$ is approximately
$1.7$ for a 95\% confidence. For instance, four additional
misclassifications out of 10,000 examples is sufficient to make such a
determination. This correspond to a difference in error rate of
$0.04\%$, roughly ten times smaller than what would be needed to
observe disjoint error bars~\eqref{eq:wald}.  
\changed{This advantage becomes very significant when 
combined with a Bonferroni-style
correction:}~$K$~pairwise comparisons 
remain simultaneously valid with confidence $1{-}\eta$ if all 
comparisons satisfy
\begin{eqnarray*}
  n_{12}-n_{21} & \geq & \sqrt2\:\:\erfcm1\!\left(\frac{2\eta}{K}\right) \:
  \sqrt{n_{12}+n_{21}} 
  %\\ K & \leq & 2\eta / \erfc\left(\frac{n_{12}-n_{21}}{\sqrt{2(n_{12}+n_{21})}}\right)~.
\end{eqnarray*}
For instance, in the realistic situation
\[
   n=10000\,,~~ n_1=200\,,~~
   n_{12}=40\,,~~ n_{21}=10\,,~~
   n_2=n_1-n_{12}+n_{21}=170\,,
\]
the conclusion that classifier~1 is worse than classifier~2 remains
valid with confidence 95\% as long as it is part of a series of 
$K{\leq}4545$ pairwise comparisons. 
In contrast, after merely $K{=}50$ experiments, the 95\% confidence 
interval for the absolute error rate of classifier~1 is already 
$2\%\pm0.5\%$, too large to distinguish it from the error rate
of classifier~2.
We should therefore expect that repeated model selection on the same
test set leads to decisions that remain valid far longer than the
corresponding absolute error rates.\footnote{See~\citep{feldman-2019}
for a different perspective on this issue.}

\begin{figure}[p]
    \centering
    \begin{tabular}{cc}
      \includegraphics[width=.48\linewidth]{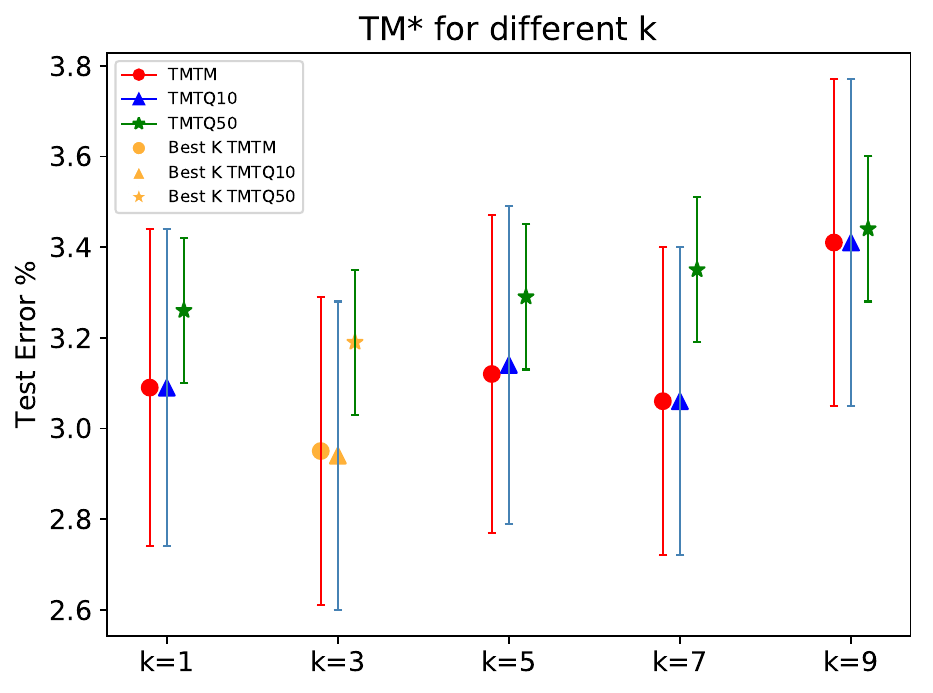} &
      \includegraphics[width=.48\linewidth]{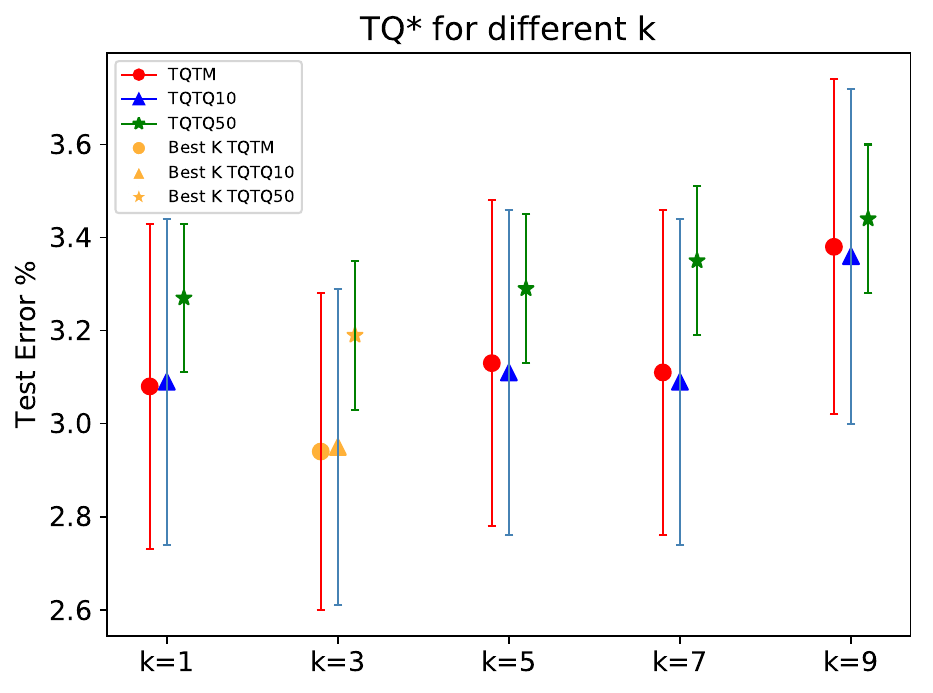} \\[-0.7ex]
      %(a) & (b) \\[-0.7ex]
    \end{tabular}
    \caption{KNN error rates for various values of $k$
    using either the MNIST (left plot) or QMNIST (right plot)
    training sets.
    Red circles: testing on MNIST. 
    Blue triangles: testing on its QMNIST counterpart. 
    Green stars: testing on the 50,000 new QMNIST testing examples.}
    \label{fig:knn}
   
    \bigskip
    
    \begin{tabular}{cc}
      \includegraphics[width=.48\linewidth]{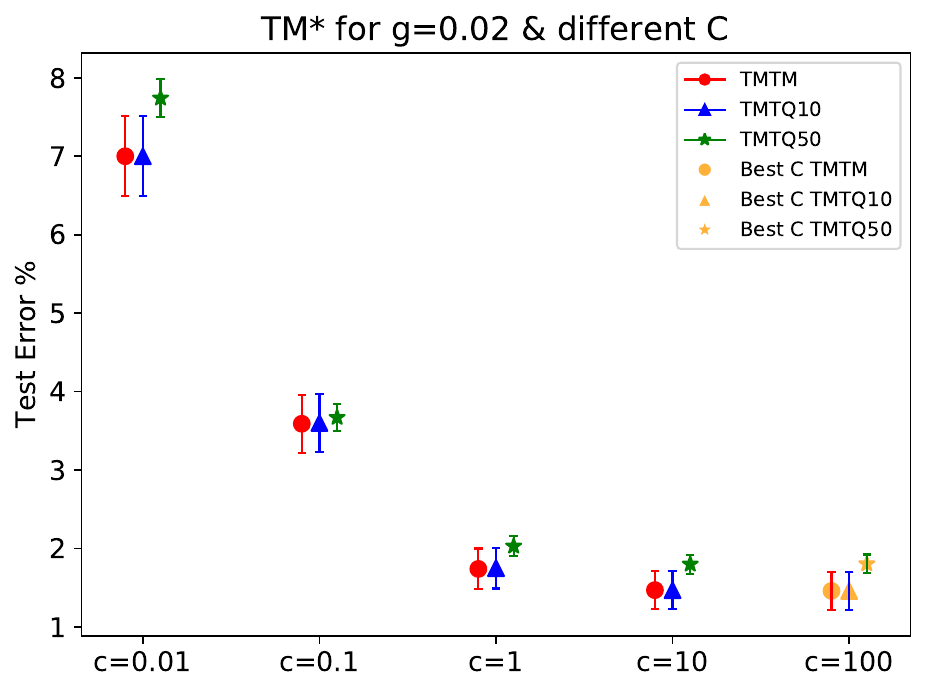} &
      \includegraphics[width=.48\linewidth]{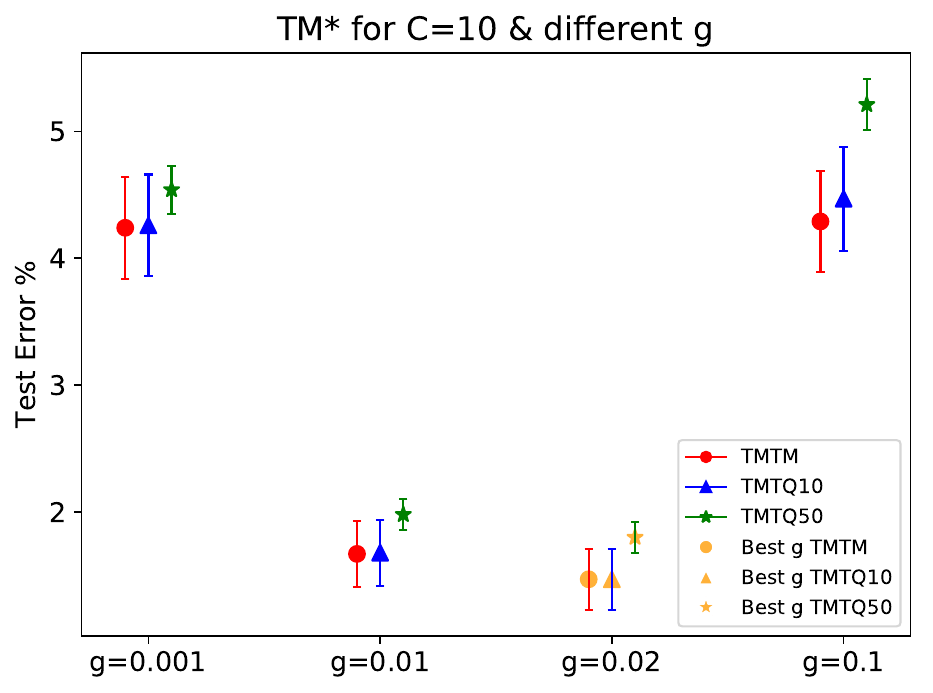} \\[-0.7ex]
      %(a) & (b)  \\[-0.7ex]
      %\includegraphics[width=.48\linewidth]{svmdiffcq} &
      %\includegraphics[width=.48\linewidth]{svmdiffgq} \\[-0.7ex]
      %(c) & (d) 
    \end{tabular}
    \caption{SVM error rates for various values of the regularization parameter $C$ (left plot) and the RBF kernel parameter $g$ (right plot) after training on the MNIST training set, using the same color and symbols as figure~\ref{fig:knn}.}
    \label{fig:svm}

    \bigskip

    \begin{tabular}{cc}
      \includegraphics[width=.48\linewidth]{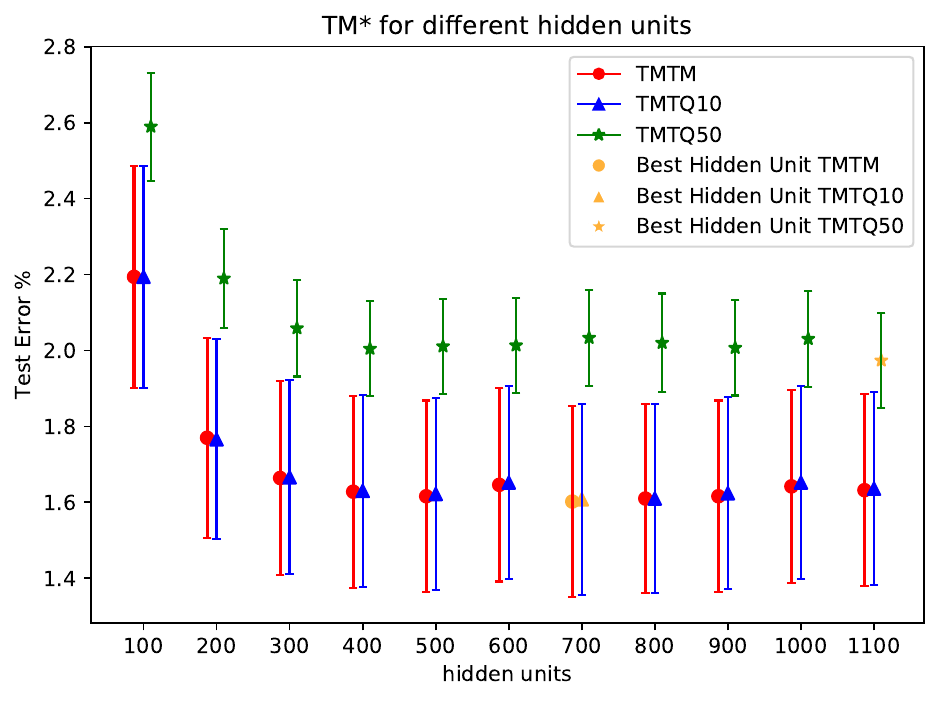} &
      \includegraphics[width=.48\linewidth]{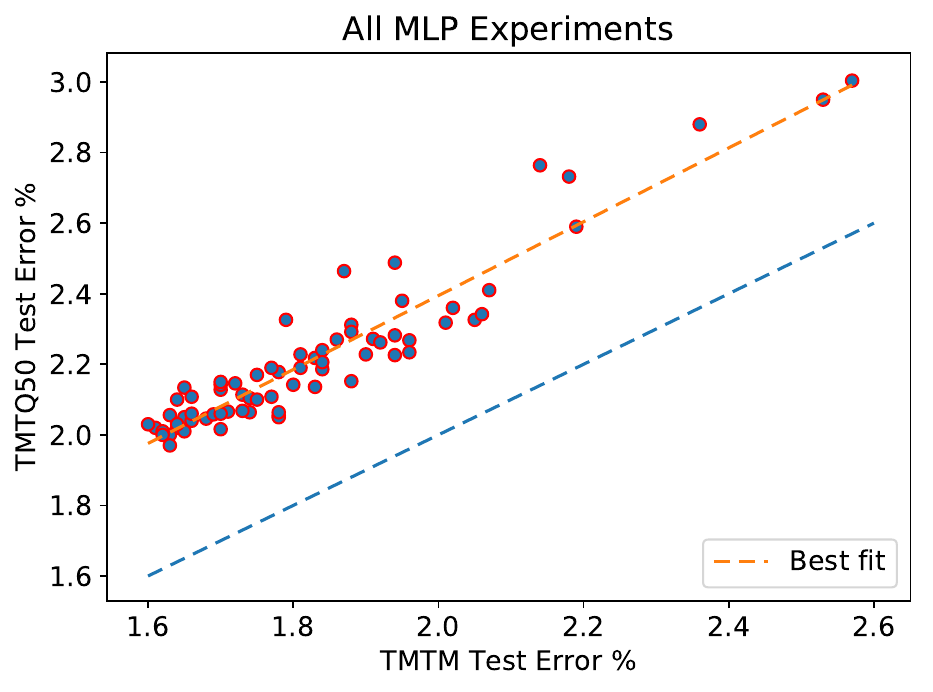} \\[-0.7ex]
      %(a) & (b)  \\[-0.7ex]
    \end{tabular}
    \caption{Left plot: MLP error rates for various hidden layer sizes after training on MNIST, using the same color and symbols as figure~\ref{fig:knn}. Right plot: scatter plot comparing the MNIST and QMNIST50K testing errors for all our MLP experiments.}
    \label{fig:mlp}
\end{figure}

\subsection{Results}
\label{sec:experiments}

We report results using two training sets, namely the MNIST training set and
the QMNIST reconstructions of the MNIST training digits, and three testing sets,
namely the official MNIST testing set with 10,000 samples (MNIST), the reconstruction of the official MNIST testing digits (QMNIST10K), and the reconstruction of the lost 50,000 testing samples (QMNIST50K). We use the names TMTM, TMTQ10, TMTQ50 to identify results measured on these three testing sets after training on the MNIST training set. Similarly we use the names TQTM, TQTQ10, and TQTQ50, for results obtained after training on the QMNIST training set and testing on the three test sets.  None of these results involves data augmentation or preprocessing steps such as deskewing, noise removal, blurring, jittering, elastic deformations, etc. 

Figure~\ref{fig:knn} (left plot) reports the testing error rates obtained with KNN for various values of the parameter~$k$ using the MNIST training set as reference points.  The QMNIST50K results are slightly worse but within the confidence intervals. The best $k$ determined on MNIST is also the best $k$ for QMNIST50K.  Figure~\ref{fig:knn} (right plot) reports similar results and conclusions when using the QMNIST training set as a reference point.

Figure~\ref{fig:svm} reports testing error rates obtained with RBF kernel SVMs after training on the MNIST training set with various
values of the hyperparameters $C$ and $g$. The QMNIST50 results
are consistently higher but still fall within the confidence intervals except maybe for mis-regularized models. Again the hyperparameters achieving the best MNIST performance also achieve the best QMNIST50K performance.

Figure~\ref{fig:mlp} (left plot) provides similar results for
a single hidden layer multilayer network with various hidden layer sizes, averaged over five runs. The QMNIST50K results again appear consistently worse than the MNIST test set results. On the one hand, the best QMNIST50K performance is achieved for a network with 1100 hidden units whereas the best MNIST testing error is achieved by a network with 700 hidden units. On the other hand, all networks with 300 to 1100 hidden units perform very similarly on both MNIST and QMNIST50, as can be seen in the plot. A 95\% confidence interval paired test on representative runs reveals no statistically significant differences between the MNIST test performances of these networks.  
Each point in figure~\ref{fig:mlp} (right plot) gives the MNIST and QMNIST50K testing error rates of one MLP experiment. This plot includes experiments with several hidden layer sizes and also several minibatch sizes and learning rates. We were only able to replicate the reported 1.6\% error rate \citet{lecun-98h} using minibatches of five or less examples.

\begin{figure}
    \centering
    \includegraphics[width=0.8\linewidth]{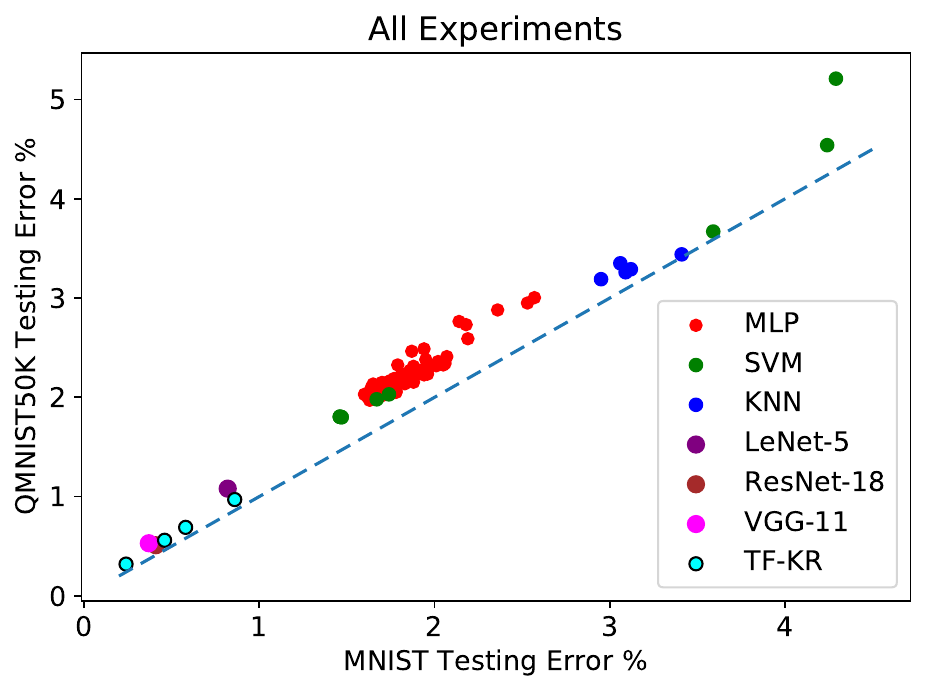}
    \caption{Scatter plot comparing the MNIST and QMNIST50K testing performance
    of all the models trained on MNIST during the course of this study.}
    \label{fig:cluster}
\end{figure}

Finally, Figure~\ref{fig:cluster} summarizes all the experiments reported above.
It also includes several flavors of convolutional networks: the Lenet5 results were
already presented in Table~\ref{tbl:lenet5}, the VGG-11 \citep{simonyan2014very} and ResNet-18 \citep{he2016deep} results are representative of the modern CNN 
architectures currently popular in computer vision. 
We also report results obtained using four models from the TF-KR MNIST challenge.\footnote{\url{https://github.com/hwalsuklee/how-far-can-we-go-with-MNIST}}
Model TFKR-a\footnote{TFKR-a: \url{https://github.com/khanrc/mnist}} is an 
ensemble two VGG- and one ResNet-like models trained with an augmented version of the MNIST training set. 
Models TFKR-b\footnote{TFKR-b: \url{https://github.com/bart99/tensorflow/tree/master/mnist}}, TFKR-c\footnote{TFKR-c: \url{https://github.com/chaeso/dnn-study}}, and 
TFKR-d\footnote{TFKR-d: \url{https://github.com/ByeongkiJeong/MostAccurableMNIST_keras}} 
are single CNN models with varied architectures.
This scatter plot shows that the QMNIST50 error rates are consistently 
slightly higher than the MNIST testing errors. However, the plot 
also shows that comparing the MNIST testing set performances of various models 
provides a near perfect ranking of the corresponding QMNIST50K performances.
In particular, the best performing model on MNIST, TFKR-a,
remains the best performing model on QMNIST50K.

% ========================================
\section{Conclusion}

We have recreated a close approximation of the MNIST preprocessing chain. 
Not only did we track each MNIST digit to its NIST source image and 
associated metadata, but also recreated the original MNIST test set, 
including the 50,000 samples that were never distributed.
These fresh testing samples allow us to investigate how
the results reported on a standard testing set suffer from repeated
experimentation. Our results confirm the trends observed 
by~\citet{recht2018cifar,pmlr-v97-recht19a}, albeit on a different dataset
and in a substantially more controlled setup.  All these results
essentially show that the ``testing set rot'' problem exists 
but is far less severe than feared. Although the repeated usage 
of the same testing set impacts absolute performance numbers,
it also delivers pairing advantages that help model selection in the long run.
\changed{In practice, this suggests that a shifting data distribution
is far more dangerous than overusing an adequately 
distributed testing set.}

% ========================================
\newpage
\ifnotanonymous
\section*{Acknowledgments}
We thank Chris Burges, Corinna Cortes, and Yann LeCun 
for the precious information they were able
to share with us about the birth of MNIST. We thank Larry Jackel for instigating the whole MNIST project and for commenting on this "cold case". We thank Maithra Raghu for pointing out how QMNIST could be used to corroborate the results of~\citet{pmlr-v97-recht19a}. We thank Ben Recht, Ludwig Schmidt and Roman Werpachowski for their constructive comments. 
\fi

\bibliographystyle{plainnat}
\bibliography{qmnist}

% =========================================

\ifsupplementary
\newpage
\appendix

\begin{center}
    \huge\textbf{Supplementary Material}
\end{center}

\vspace{4ex}

This section provides additional tables and plots.
\bigskip
\FloatBarrier
\begin{table}[ht]
  %%% should we move this one to the end?
  \centering
  \caption{\label{tbl:KNN3} Misclassification rates of the best KNN model obtained when $k$ is set to 3.
    Model trained on both the MNIST and QMNIST
    training sets and tested on the MNIST test set, and the two
QMNIST test sets of size 10,000 \& 50,000 each.}
  \smallskip
  \begin{tabular}{lccc}
    \toprule
    \bf Test on & \bf MNIST & \bf QMNIST10K & \bf QMNIST50K \\
    \midrule
    Train on MNIST & $2.95\%$ ($\pm 0.34\%$) & $2.94\%$ ($\pm 0.34\%$) & $3.19\%$ ($\pm 0.16\%$)\\
    Train on QMNIST & $2.94\%$ ($\pm 0.34\%$) & $2.95\%$ ($\pm 0.34\%$) & $3.19\%$ ($\pm 0.16\%$) \\
    \midrule
  \end{tabular}
\end{table}
\FloatBarrier

\FloatBarrier
\begin{table}[ht]
  \centering
  \caption{\label{tb2:svm} Misclassification rates of a SVM when hyperparameters $C$ = 10 \& $g$ = 0.02. Training and testing schemes are similar to Table \ref{tbl:KNN3}.}
  \smallskip
  \begin{tabular}{lccc}
    \toprule
    \bf Test on & \bf MNIST & \bf QMNIST10K & \bf QMNIST50K \\
    \midrule
    Train on MNIST & $1.47\%$ ($\pm 0.24\%$) & $1.47\%$ ($\pm 0.24\%$) & $1.8\%$ ($\pm 0.12\%$)\\
    Train on QMNIST & $1.47\%$ ($\pm 0.24\%$) & $1.48\%$ ($\pm 0.24\%$) & $1.8\%$ ($\pm 0.12\%$) \\
    \midrule
  \end{tabular}
\end{table}
\FloatBarrier
\begin{table}[ht]
  \centering
  \caption{\label{tb3:lenet5v2} Misclassification rates of an MLP with a 800 unit hidden layer. Training and testing schemes are similar to Table \ref{tbl:KNN3}.}
  \smallskip
  \begin{tabular}{lccc}
    \toprule
    \bf Test on & \bf MNIST & \bf QMNIST10K & \bf QMNIST50K \\
    \midrule
    Train on MNIST & $1.61\%$ ($\pm 0.25\%$) & $1.61\%$ ($\pm 0.25\%$) & $2.02\%$ ($\pm 0.13\%$)\\
    Train on QMNIST & $1.63\%$ ($\pm 0.25\%$) & $1.63\%$ ($\pm 0.25\%$) & $2\%$ ($\pm 0.13\%$) \\
    \midrule
  \end{tabular}
\end{table}
\FloatBarrier

\begin{table}[ht]
  \centering
  \caption{\label{tb4:vgg} Misclassification rates of a VGG-11 model. Training and testing schemes are similar to Table \ref{tbl:KNN3}.}
  \smallskip
  \begin{tabular}{lccc}
    \toprule
    \bf Test on & \bf MNIST & \bf QMNIST10K & \bf QMNIST50K \\
    \midrule
    Train on MNIST & $0.37\%$ ($\pm 0.12\%$) & $0.37\%$ ($\pm 0.12\%$) & $0.53\%$ ($\pm 0.06\%$)\\
    Train on QMNIST & $0.39\%$ ($\pm 0.12\%$) & $0.39\%$ ($\pm 0.12\%$) & $0.53\%$ ($\pm 0.06\%$) \\
    \midrule
  \end{tabular}
\end{table}

\FloatBarrier

\begin{table}[ht]
  \centering
  \caption{\label{tb5:resnet} Misclassification rates of a ResNet-18 model. Training and testing schemes are similar to Table \ref{tbl:KNN3}.}
  \smallskip
  \begin{tabular}{lccc}
    \toprule
    \bf Test on & \bf MNIST & \bf QMNIST10K & \bf QMNIST50K \\
    \midrule
    Train on MNIST & $0.41\%$ ($\pm 0.13\%$) & $0.42\%$ ($\pm 0.13\%$) & $0.51\%$ ($\pm 0.06\%$)\\
    Train on QMNIST & $0.43\%$ ($\pm 0.13\%$) & $0.43\%$ ($\pm 0.13\%$) & $0.50\%$ ($\pm 0.06\%$) \\
    \midrule
  \end{tabular}
\end{table}

\begin{table}[ht]
  \centering
  \caption{\label{tb6:kaggle} Misclassification rates of top TF-KR MNIST models trained on the MNIST training se and tested on the MNIST, QMNIST10K and QMNIST50K testing sets. }
  \smallskip
  \begin{tabular}{lccc}
    \toprule
    \bf Github Link & \bf MNIST & \bf QMNIST10K & \bf QMNIST50K \\
    \midrule
    TFKR-a & $0.24\%$ ($\pm 0.10\%$) & $0.24\%$ ($\pm 0.10\%$) & $0.32\%$ ($\pm 0.05\%$)\\
    TFKR-b & $0.86\%$ ($\pm 0.18\%$) & $0.86\%$ ($\pm 0.18\%$) & $0.97\%$ ($\pm 0.09\%$) \\
    TFKR-c & $0.46\%$ ($\pm 0.14\%$) & $0.47\%$ ($\pm 0.14\%$) & $0.56\%$ ($\pm 0.07\%$) \\
    TFKR-d & $0.58\%$ ($\pm 0.15\%$) & $0.58\%$ ($\pm 0.15\%$) & $0.69\%$ ($\pm 0.07\%$) \\
    \midrule
  \end{tabular}
\end{table}

\FloatBarrier

\begin{figure}[p]
    \centering
    \begin{tabular}{cc}
      \includegraphics[width=.48\linewidth]{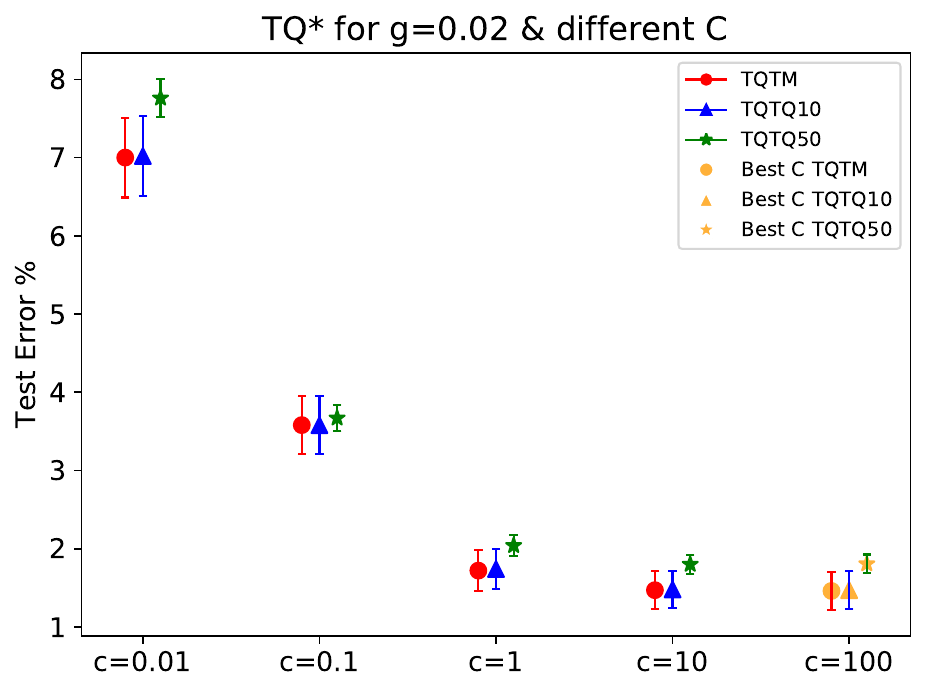} &
      \includegraphics[width=.48\linewidth]{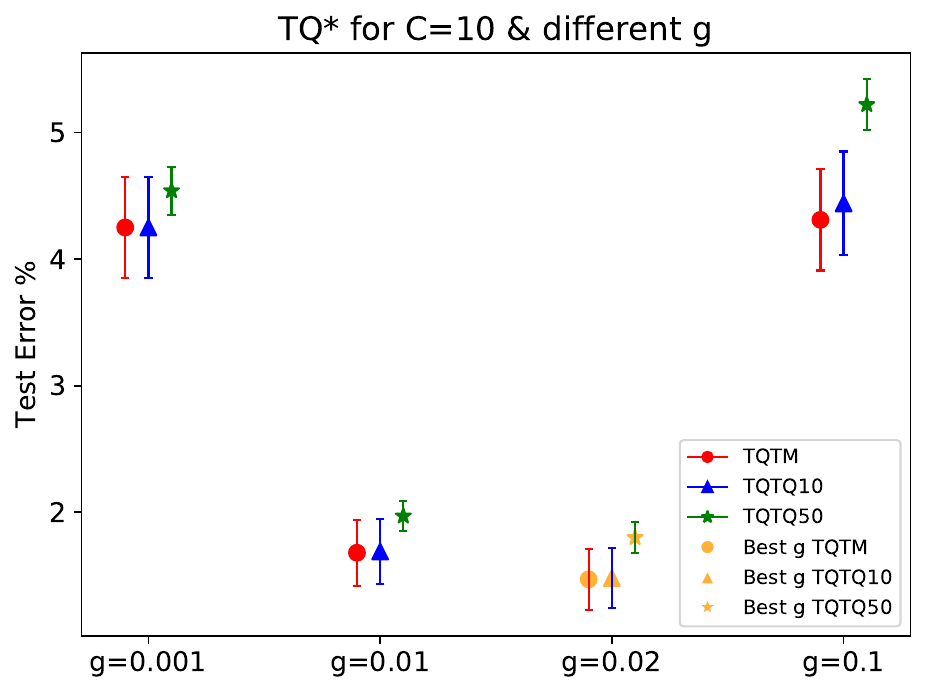} \\[-0.7ex]
    \end{tabular}
    \caption{SVM error rates for various values of the regularization parameter $C$ (left plot) and the RBF kernel parameter $g$ (right plot) after training on the QMNIST training set. Red circles: testing on MNIST. 
    Blue triangles: testing on its QMNIST counterpart. 
    Green stars: testing on the 50,000 new QMNIST testing examples.}
    \label{fig:svmq}
   
    \bigskip
    
    \begin{tabular}{cc}
      \includegraphics[width=.48\linewidth]{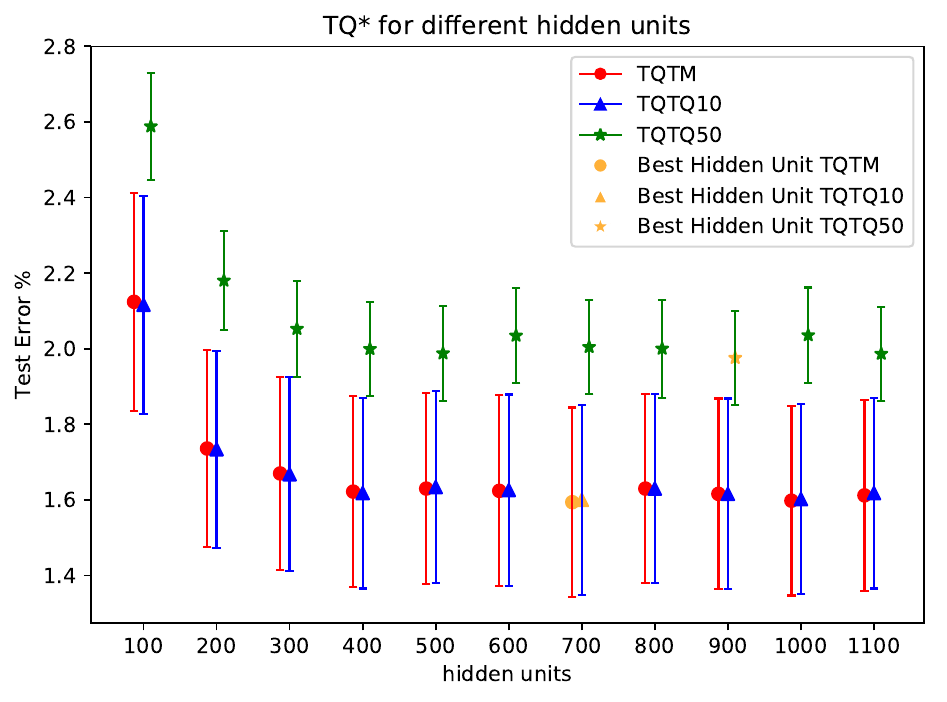} &
      \includegraphics[width=.48\linewidth]{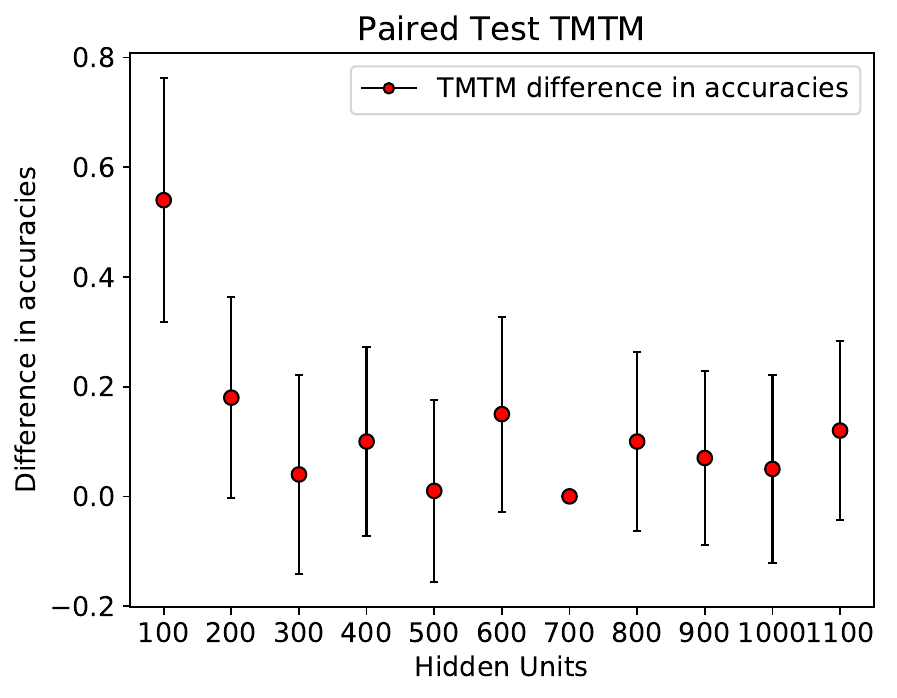} \\[-0.7ex]
    \end{tabular}
    \caption{Left plot : MLP error rates for various hidden layer sizes after training on the QMNIST training set, using the same testing scheme as figure~\ref{fig:svmq}.
    Right plot: Paired test of MLPs with different hidden layer sizes and MLP with 700 hidden units (which performs best on MNIST test set). All of the MLPs used in this plot were trained and tested on MNIST.}
    \label{fig:mlpleft}

    \bigskip

    \begin{tabular}{cc}
      \includegraphics[width=.48\linewidth]{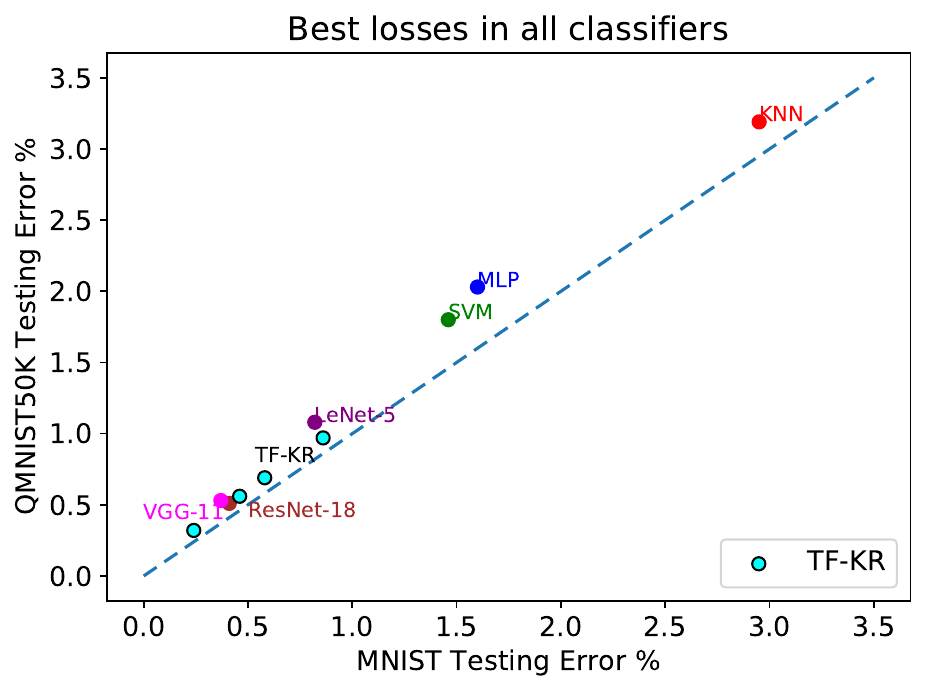} \\[-0.7ex]
      
    \end{tabular}
    \caption{Scatter plot comparing the best MNIST and QMNIST50K testing performance of all the classifiers trained on MNIST during the course of this study.}
    \label{fig:allleft}
\end{figure}
\fi  %%%%%%%%% ifsupplementary

% =========================================

\end{document}